%% file: main.tex
\documentclass[conference]{IEEEtran}
\IEEEoverridecommandlockouts
\usepackage{graphicx}
\usepackage{cite}
\usepackage{amsmath,amssymb,amsfonts}
\usepackage{algorithmic}
\usepackage{textcomp}
\usepackage{xcolor}
\usepackage{booktabs}
\usepackage{makecell}
\usepackage{array}
\usepackage{multirow}
\usepackage{verbatim}
\usepackage[T1]{fontenc}
\usepackage{float}
\usepackage{svg}
\usepackage{colortbl}
\usepackage{subfigure}
\usepackage{url}
\usepackage{subcaption}
\def\BibTeX{{\rm B\kern-.05em{\sc i\kern-.025em b}\kern-.08em
    T\kern-.1667em\lower.7ex\hbox{E}\kern-.125emX}}
\begin{document}
\title{Multitask LSTM for Arboviral Outbreak Prediction Using Public Health Data}
\author{
\IEEEauthorblockN{Lucas R. C. Farias}
\IEEEauthorblockA{Universidade Católica de Pernambuco\\
Universidade Federal de Pernambuco\\
CESAR School\\
Recife, Brazil\\
lucas.farias@unicap.br}
\and
\IEEEauthorblockN{Talita P. Silva}
\IEEEauthorblockA{Universidade Católica de Pernambuco\\
 Recife, Brazil \\
talita.2021104938@unicap.br}
\and
\IEEEauthorblockN{Pedro H. M. Araújo}
\IEEEauthorblockA{Universidade Católica de Pernambuco\\
Universidade de Pernambuco\\
 Recife, Brazil \\
pedro.araujo@unicap.br}
}
\IEEEoverridecommandlockouts
\maketitle
\IEEEpubidadjcol

\begin{abstract}

This paper presents a multitask learning approach based on long-short-term memory (LSTM) networks for the joint prediction of arboviral outbreaks and case counts of dengue, chikungunya, and Zika in Recife, Brazil. Leveraging historical public health data from DataSUS (2017–2023), the proposed model concurrently performs binary classification (outbreak detection) and regression (case forecasting) tasks. A sliding window strategy was adopted to construct temporal features using varying input lengths (60, 90, and 120 days), with hyperparameter optimization carried out using Keras Tuner. Model evaluation used time series cross-validation for robustness and a held-out test from 2023 for generalization assessment. The results show that longer windows improve dengue regression accuracy, while classification performance peaked at intermediate windows, suggesting an optimal trade-off between sequence length and generalization. The multitask architecture delivers competitive performance across diseases and tasks, demonstrating the feasibility and advantages of unified modeling strategies for scalable epidemic forecasting in data-limited public health scenarios.
\end{abstract}


\input{section/introduction}
\input{section/methodology}
\input{section/experimental_setup}

\input{section/results}
\input{section/conclusion}

\bibliographystyle{ieeetr} 
\bibliography{bibliography.bib}

\end{document}

%% file: section/introduction.tex
\section{Introduction}

Arboviral diseases such as dengue, chikungunya, and Zika continue to pose significant public health challenges in Brazil, particularly in the Northeast region, which experiences recurrent and severe outbreaks~\cite{moyo2023emerging, eryando2022spatial}. Notably, in early 2024, the state of Pernambuco reported a substantial surge in dengue incidence, with an average epidemic percent change (AEPC) of 55\% (95\% Confidence Interval: 43.4--67.4; \emph{p}~<~0.001)~\cite{souza2024space}.

Time series modeling has emerged as a valuable tool for anticipating outbreak trends, enabling timely public health interventions. Prior studies have demonstrated the effectiveness of models built solely on normalized historical incidence data~\cite{Dama2021}. Among data-driven methods, deep learning approaches, particularly long short-term memory (LSTM) networks, have consistently outperformed traditional models in modeling temporal dependencies within epidemiological time series~\cite{jia2019predicting, make5010013, Absar2022, Darwish2020}.

Recent literature consistently indicates that machine learning models, especially neural networks and LSTM-based architectures, outperform traditional statistical approaches in infectious disease forecasting. Jia et al.~\cite{jia2019predicting} and Santangelo et al.~\cite{make5010013} reported improved predictive performance from recurrent neural networks and ensemble methods when compared to regression-based and compartmental models such as SIR. Nevertheless, limitations persist, particularly concerning data quality, preprocessing demands, and overfitting risks, all of which may hinder model generalizability. Absar et al.~\cite{Absar2022} further validated the strength of LSTM models by accurately forecasting COVID-19 trends, reinforcing their relevance for real-time public health planning.

LSTM networks have consistently surpassed both statistical and machine learning baselines in a variety of epidemiological contexts. For example, Darwish, Rahhal, and Jafar~\cite{Darwish2020} demonstrated that LSTM models outperformed Naïve, Drift, TBATS, Generalized Linear Models, Support Vector Regression, and Random Forests in predicting influenza-like illness in Syria, highlighting their adaptability and robustness in data-limited scenarios.

Motivated by these findings, this paper introduces a multitask learning framework based on LSTM networks that simultaneously performs classification (outbreak detection) and regression (case forecasting) for dengue, chikungunya, and Zika. The proposed model leverages historical public health data from DataSUS and addresses key modeling challenges through shared temporal representations. Recife, Brazil, was selected as the study site due to its epidemiological importance and socioeconomic vulnerability~\cite{souza2024space}, to improve surveillance and inform targeted health interventions.

The main contributions of this paper are as follows: 

\begin{enumerate}
    \item A multitask LSTM model for arboviral outbreak prediction and case forecasting in Recife;
    \item An empirical evaluation of different temporal window sizes on model performance;
    \item A practical application of machine learning to open government health data (DataSUS).
\end{enumerate}

The remainder of this paper is organized as follows: Section II presents the dataset and model; Section III describes the experimental setup; Section IV discusses the results; and Section V concludes the paper with future directions.

%% file: section/methodology.tex
\section{Methods}\label{methods}

\begin{figure*}[htbp]
\centering

\begin{minipage}{0.32\textwidth} 
    \centering
    \includegraphics[width=\textwidth]{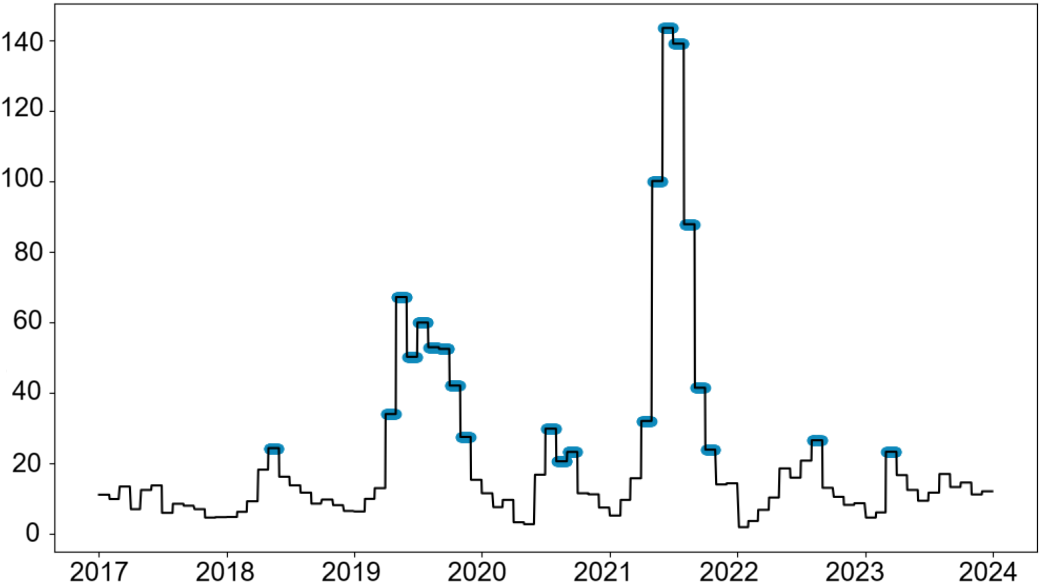}
    \caption*{(a) Dengue}
    \label{fig:inc_dengue}
\end{minipage}%
\hfill 
\begin{minipage}{0.32\textwidth}
    \centering
    \includegraphics[width=\textwidth]{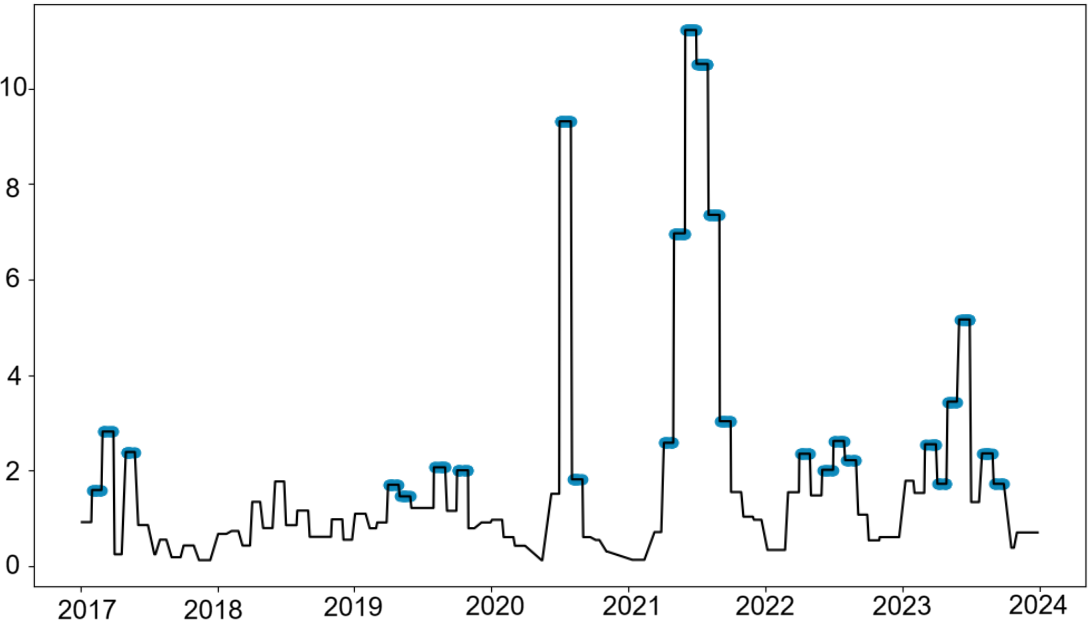}
    \caption*{(b) Zika}
    \label{fig:inc_zika}
\end{minipage}%
\hfill 
\begin{minipage}{0.32\textwidth}
    \centering
    \includegraphics[width=\textwidth]{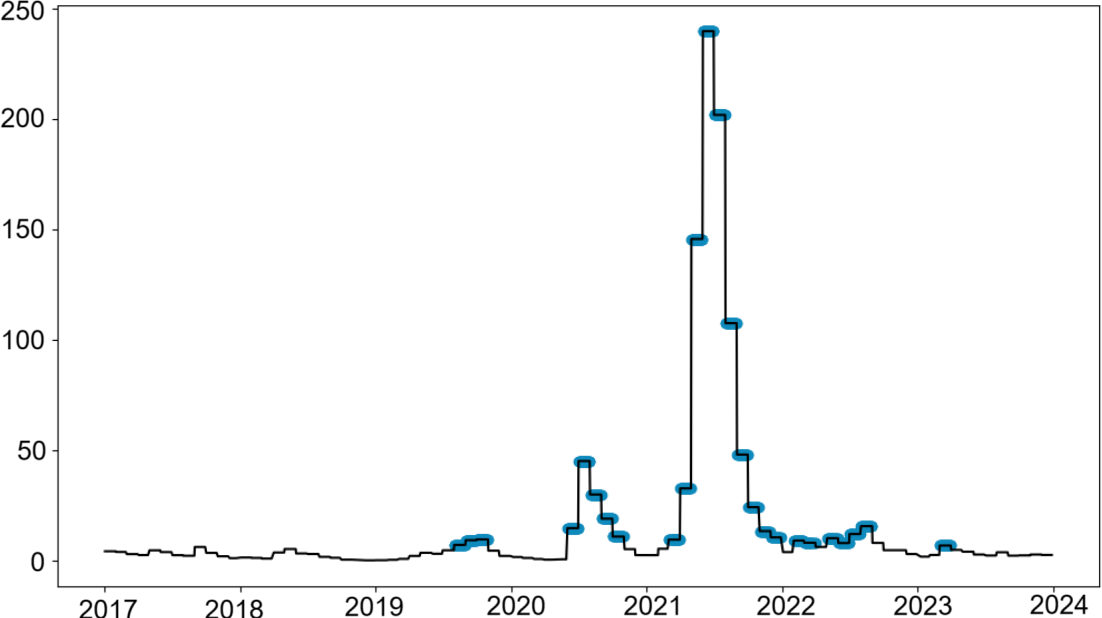}
    \caption*{(c) Chikungunya}
    \label{fig:inc_chik}
\end{minipage}

\caption{Monthly incidence rates of arboviral diseases in Recife, Brazil, from 2017 to 2024. Each subplot presents the incidence rate per 100,000 inhabitants for (a) dengue, (b) Zika, and (c) chikungunya. The solid black lines represent the temporal progression of monthly incidence, while the blue markers indicate months identified as outbreak periods, defined as values exceeding the disease-specific threshold percentile.}
\label{fig:outbreak}
\end{figure*}

\subsection{Dataset}

We used epidemiological data from the Brazilian public health system (DataSUS) ~\cite{datasus}, which provides records from the Notifiable Diseases Information System (SINAN) \cite{sinan}. SINAN includes structured data on 49 diseases, encompassing 108 attributes such as patient demographics (e.g., age, sex, race) and case evolution (e.g., symptom onset, diagnosis, case evolution). Access and preprocessing were performed using the PySUS library \cite{pysus}.

Our analysis focused on confirmed cases of dengue, chikungunya, and Zika in Recife from 2017 to 2023. Data preprocessing steps involved filtering by municipality, handling missing values, and calculating incidence rates standardized per 100,000 inhabitants. These operations ensured data integrity and facilitated consistent modeling.

\subsection{Long Short-Term Memory Networks}

Recurrent Neural Networks (RNNs) are designed to model temporal sequences, but suffer from vanishing gradients when learning long-range dependencies. Long Short-Term Memory (LSTM) networks address this by incorporating gated mechanisms (input, forget, and output gates) that control information flow, making them effective for time series forecasting in epidemiology.

\subsection{Incidence and Outbreak Definition}

We computed monthly incidence rates using population estimates from the Brazilian Institute of Geography and Statistics (IBGE) \cite{ibge_censo_2010, ibge_estimativas_dou}, accessed through Datapedia \cite{datapedia}. When annual estimates were missing, linear interpolation was applied using a first-order approximation \cite{shryock2004methods}. The monthly incidence rate $TI_{monthly}$ was computed as:
\begin{equation}
TI_{monthly} = \frac{N_{cases}}{P} \times 100.000
\label{eq:taxa_inc}
\end{equation}
where $N_{cases}$ represents the number of confirmed cases of a given disease in a specific month, and $P$ denotes the estimated population of the region in that same month. The resulting metric expresses the number of new cases per 100,000 inhabitants, allowing for standardized comparisons across time and disease types.

To define outbreak periods, disease-specific thresholds were derived from the empirical distribution of monthly incidence rates. We adopted the 75th percentile as the cutoff for dengue and the 70th percentile for chikungunya and Zika. The higher threshold for dengue accounts for its higher baseline incidence, reducing the risk of false positives, whereas lower thresholds for the other diseases increase sensitivity to abnormal spikes.

Each month was subsequently labeled as:
\begin{itemize}
    \item \textbf{1 (outbreak):} if $TI_{monthly}$ exceeded the disease-specific threshold.
    \item \textbf{0 (non-outbreak):} otherwise.
\end{itemize}

Outbreak detection was formulated as a binary classification task, while incidence prediction was framed as a regression problem. Both were integrated into a multitask LSTM framework capable of learning from shared temporal dependencies. This joint formulation aims to enhance generalization and improve the ability of the model to identify early signals of disease escalation. Figure~\ref{fig:outbreak} illustrates the temporal dynamics of incidence rates and the corresponding outbreak labels.

\begin{figure*}[ht]
     \centering
    \includegraphics[width=1\linewidth]{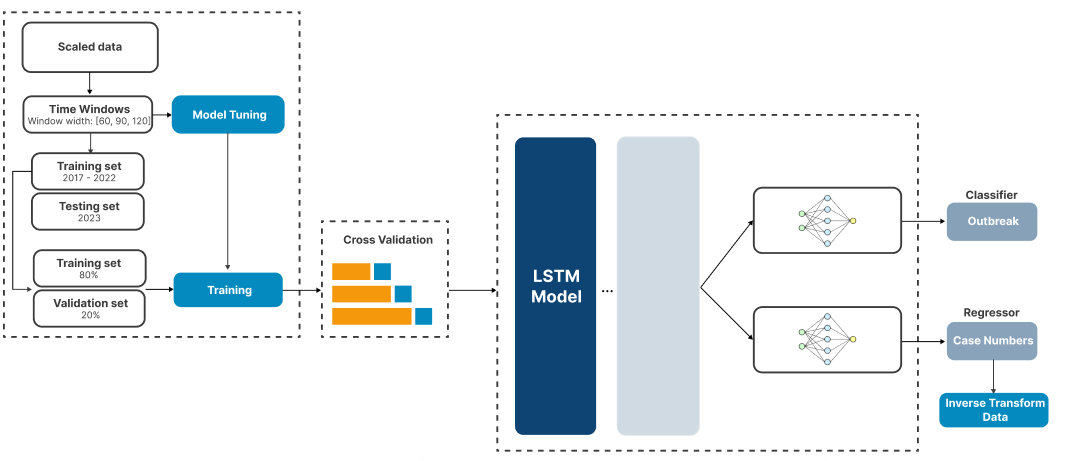}
    \caption{Workflow of the proposed multitask LSTM-based prediction framework. The pipeline begins with scaled epidemiological time series data, from which overlapping input windows (60, 90, and 120 days) are generated. The dataset is divided into training (2017–2022), validation (20\% of training), and testing (2023) subsets. Model tuning is performed using Keras Tuner before training with time series cross-validation. The multitask LSTM network simultaneously outputs (i) a binary classifier for outbreak detection and (ii) a regressor for case count forecasting. Predicted values are post-processed via inverse normalization to restore interpretability.}
     \label{fig:VisaoGeral}
\end{figure*}

\subsection{Preprocessing and Feature Engineering}

All numerical features were normalized using Min-Max scaling:
\begin{equation} X_{\text{norm}} = \frac{X - X_{\text{min}}}{X_{\text{max}} - X_{\text{min}}} \end{equation}
where $X$ is the original value of the variable. $X_{\text{min}}$ and $X_{\text{max}}$ are the minimum and maximum values of the variable, respectively.

A sliding window technique was adopted to generate temporal features from prior observations. We evaluated three input window sizes (60, 90, 120 days), defined as:
\begin{equation} X_{\text{window}} = {X_{t-1}, X_{t-2}, \dots, X_{t-T}} \end{equation}
where $X_{\text{window}}$ denotes the input vector comprising time series observations from previous time steps, $X_t$ is the value of the variable at the current time $t$, and $T$ represents the number of past observations.

\subsection{Multitask Architecture}

We implemented a multitask LSTM architecture capable of simultaneously predicting (i) outbreak probability and (ii) case counts. This approach allows the model to exploit shared temporal patterns across tasks, improving generalization and enabling joint optimization.

\subsection{Evaluation Metrics and Statistical Analysis}

To assess the performance of the multitask LSTM model, we used task-specific evaluation metrics for classification and regression. The architecture jointly performed outbreak detection and case forecasting, necessitating distinct criteria for each task.

For classification, we computed the F1-score and the area under the receiver operating characteristic curve (AUC-ROC). The F1-score, defined as the harmonic mean of precision and recall, is particularly suitable for imbalanced datasets, capturing the trade-off between false positives and false negatives~\cite{Geron2019}. Precision indicates the proportion of predicted outbreaks that were correct, while recall measures the proportion of actual outbreaks that were correctly identified.

The ROC curve provides a graphical assessment of classification performance by plotting the true positive rate against the false positive rate across varying decision thresholds. The AUC-ROC summarizes the capacity of the model to discriminate between outbreak and non-outbreak periods..

For regression, we adopted two complementary metrics: the Mean Absolute Percentage Error (MAPE) and the Median Absolute Percentage Error (MedAPE). MAPE captures the average relative deviation between predicted and observed values, while MedAPE mitigates the influence of outliers by computing the median of absolute percentage errors, enhancing robustness in datasets with extreme values.

To estimate the reliability of classification results, we applied bootstrap resampling \cite{efron1979bootstrap} with 1,000 iterations on the test set. This nonparametric method approximates the sampling distribution of performance metrics through resampling with replacement, avoiding assumptions about data distribution. The F1-score and AUC-ROC were computed for each resample, enabling the construction of 95\% confidence intervals \cite{wasserman2004all} to quantify uncertainty and support a statistically grounded evaluation of model performance.

\begin{figure*}[ht]
\centering

\begin{minipage}{0.32\textwidth} 
    \centering
    \includegraphics[width=\textwidth]{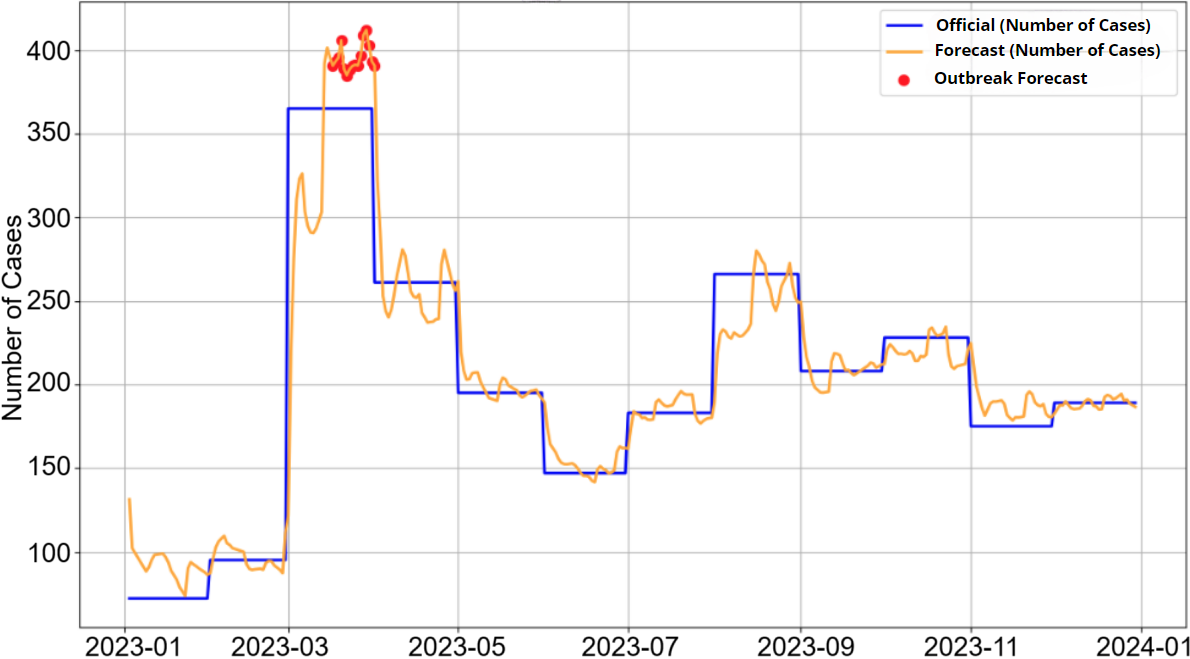}
    \caption*{(a) Dengue}
    \label{fig:dengue_lstm_simples_120_prediction}
\end{minipage}%
\hfill 
\begin{minipage}{0.32\textwidth}
    \centering
    \includegraphics[width=\textwidth]{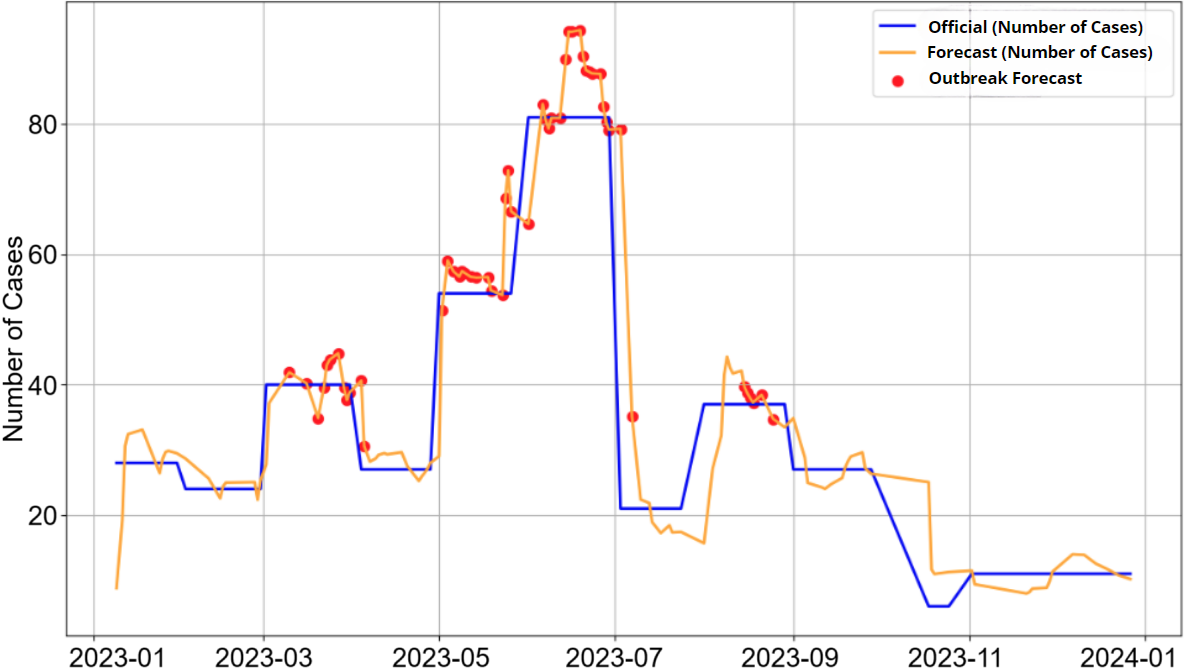}
    \caption*{(b) Zika}
    \label{fig:zika_lstm_simples_120_prediction}
\end{minipage}%
\hfill 
\begin{minipage}{0.32\textwidth}
    \centering
    \includegraphics[width=\textwidth]{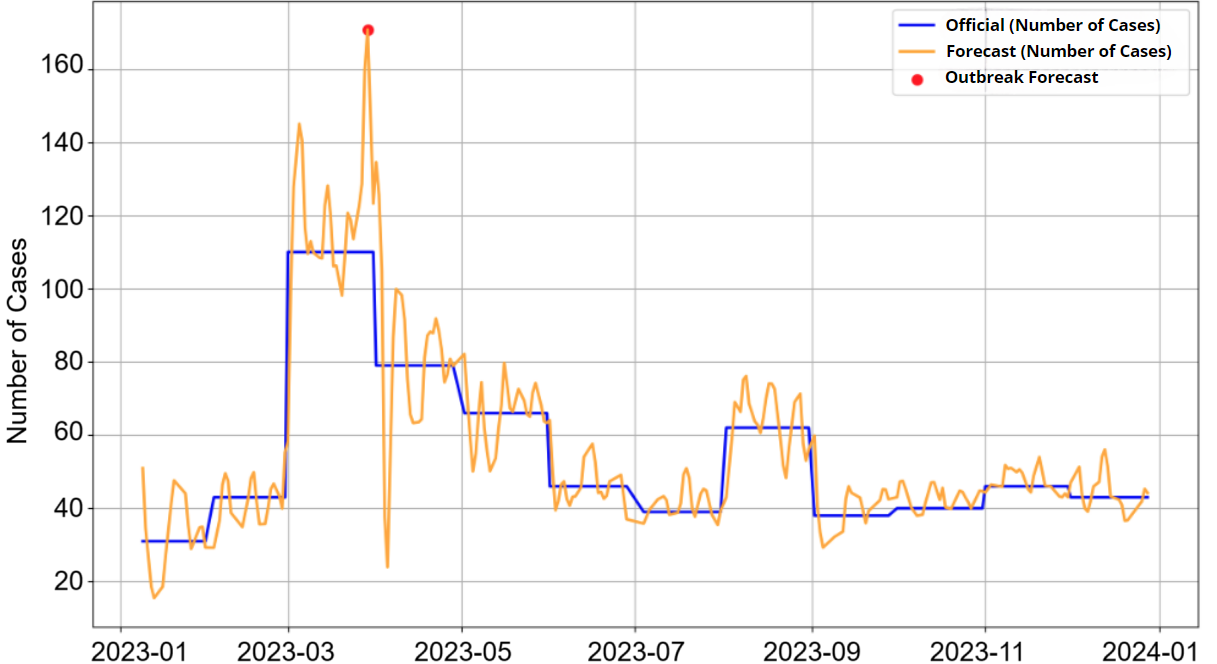}
    \caption*{(c) Chikungunya}
    \label{fig:chikungunya_lstm_simples_90_prediction}
\end{minipage}

\caption{Forecasting performance of the multitask LSTM model on the test set for three arboviral diseases: (a) dengue (120-step input), (b) Zika (120-step input), and (c) chikungunya (90-step input). In each subplot, the blue line represents the official reported case counts, the yellow line corresponds to the predicted values of the model, and red dots denote months classified as outbreak periods. The selected configurations correspond to the lowest median absolute percentage error (MedAPE) reported in Table \ref{table:forecast_erros}. All results are derived from the Simple LSTM architecture.}
\label{fig:Prediction}
\end{figure*}

\begin{figure*}[htbp]
\centering
\subfigure[Dengue\label{fig:ic_dengue}]{
\includegraphics[width=.3\textwidth]{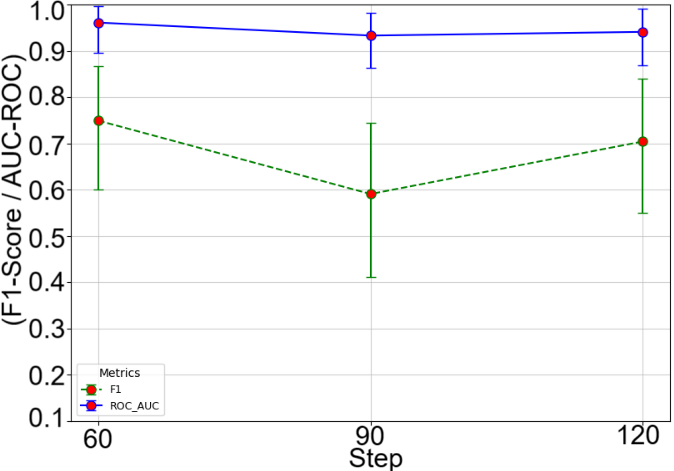}
} 
\subfigure[Zika\label{fig:ic_zika}]{
\includegraphics[width=.3\textwidth]{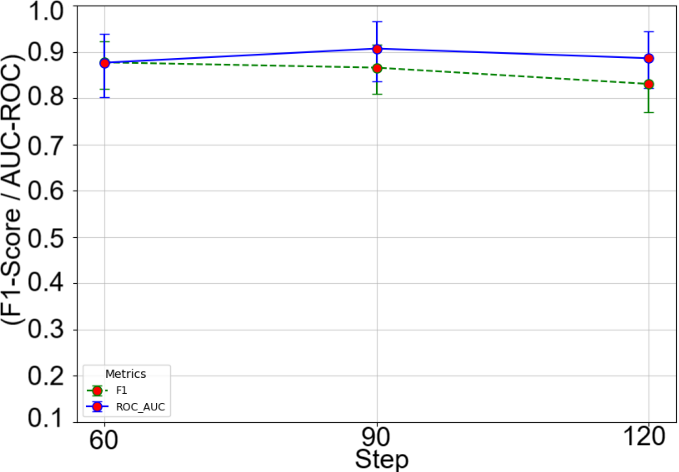}
} 
\subfigure[Chikungunya\label{fig:ic_chikungunya}]{
\includegraphics[width=.3\textwidth]{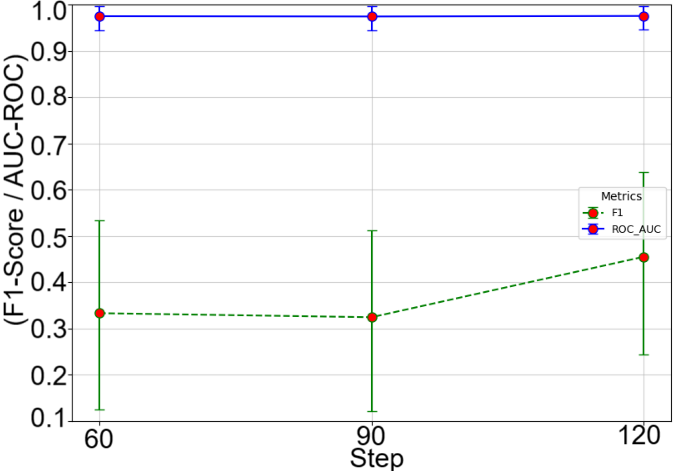}
} 
\caption{Bootstrap-based 95\% confidence intervals for the F1-score (dashed lines) and AUC-ROC (solid lines) across different input window sizes (60, 90, and 120 steps) using the Simple LSTM model. Results are presented separately for (a) dengue, (b) Zika, and (c) chikungunya. Each point represents the mean performance on the held-out test set, while vertical bars indicate the uncertainty derived from 1,000 bootstrap resamplings. }
\label{fig:95CI}
\end{figure*}

%% file: section/experimental_setup.tex
\section{Experimental Setup}

\subsection{Model Configuration and Training}

The LSTM models were developed using Python 3.10.12 and implemented with Keras 3.5.0 on a TensorFlow 2.17.1 backend. Experiments were executed on an NVIDIA Tesla T4 GPU (CUDA 12.2, driver 535.104.05). Two architectures were evaluated: a Simple LSTM with two stacked layers and a Bidirectional LSTM with three layers, including two dense layers for increased representational capacity.

To prevent overfitting, dropout regularization was applied~\cite{labach2019surveydropoutmethodsdeep}. Dense layers were included to enhance the ability of the network to model complex functions~\cite{10.1007/978-3-642-24412-4_3}. Although deeper models can improve learning, the number of layers was limited, taking advantage of the universal approximation theorem~\cite{ISMAILOV2014963} to balance model complexity and generalization capacity.

The Bidirectional LSTM enables the model to learn dependencies in both temporal directions~\cite{schuster1997bidirectional}. The multitask loss function combined binary cross-entropy (for outbreak classification) and mean squared error (for case forecasting), with equal weighting to ensure balanced optimization. Optimization was performed using the Adam algorithm~\cite{kingma2017adammethodstochasticoptimization}.

\subsection{Hyperparameter Optimization and Validation}

Model tuning was conducted using Keras Tuner~\cite{omalley2019kerastuner} with the RandomSearch strategy. The number of units in each layer was optimized for each input window size (60, 90, and 120 days). Thirty configurations were evaluated per window size, each trained for 50 epochs. Unit counts varied from 128 to 512, and the objective was to minimize validation loss.

Following hyperparameter tuning, the final models were trained using five-fold time series cross-validation over 100 epochs. Early stopping (patience = 10) and learning rate reduction on plateau~\cite{keras_reduce_lr} were used to mitigate overfitting and adapt learning rates dynamically.

The training set spanned 2017-2022, with an 80/20 split for training and validation. Data from 2023 was held out for final testing, ensuring that test samples remained completely unseen throughout the model development process. The overall pipeline is illustrated in Figure~\ref{fig:VisaoGeral}.

\subsection{Forecasting Procedure}

Final models were applied to generate forecasts for each disease using the best-performing input configuration. At each prediction step $t$, a sliding input window of the previous 60 days was used to forecast the next day’s values. The window advanced one day at a time, incorporating observed or predicted values as appropriate. Model outputs were generated in normalized scale and post-processed via inverse normalization using min-max parameters from the training set.

This rolling prediction scheme enabled temporal evaluation of the model under realistic forecasting conditions. Figure~\ref{fig:Prediction} displays examples of forecast outputs for the three diseases, and Figure~\ref{fig:95CI} presents the bootstrap-based confidence intervals for classification metrics.

\begin{table*}[ht]
\centering
\caption{Forecasting errors for dengue, chikungunya, and Zika case predictions using Simple and Bidirectional LSTM models across three input window sizes (60, 90, and 120 steps). Metrics reported include the Median Absolute Percentage Error (MedAPE) and Mean Absolute Percentage Error (MAPE). Best results per disease and metric are highlighted in bold.}
\begin{tabular}{@{}lccc|ccc@{}}
\toprule
\multirow{2}{*}{} & \multicolumn{3}{c}{\textbf{Simple LSTM}} & \multicolumn{3}{c}{\textbf{Bidirectional LSTM}} \\ 
\cmidrule(r){2-4} \cmidrule(r){5-7}
                  & 60 steps & 90 steps & 120 steps & 60 steps & 90 steps & 120 steps \\ 
\midrule
 \multicolumn{7}{l}{\textbf{MedAPE (\%)}} \\
Dengue     &7.7     &5.7     & \textbf{4.45}      & 8.0     & 6.2     & 9.13     \\
Chikungunya    & 12.5   & 9.8    & 10.7      & 17.6     &  14.2     &  \textbf{8.5}      \\
Zika      & 18.5     & 17.7     & \textbf{7.5}      &  19.0     & 19.0    & 25.0     \\
\midrule
\multicolumn{7}{l}{\textbf{MAPE (\%)}} \\
Dengue       & 10.8     & 8.5     & \textbf{7.5}      & 12.8     & 10.8     & 13.5      \\
Chikungunya       & 15.2     & \textbf{12.9}     &  13.2      &  19.6     & 18.0     & 13.4      \\
Zika      & 31.8     & 27.6    & \textbf{17.2}      & 26.3     & 26.3     & 35.0     \\
\bottomrule
\end{tabular}
\label{table:forecast_erros}
\end{table*}

\begin{table*}[ht]
\centering
\caption{Validation and testing performance of the Simple LSTM model across input window sizes of 60, 90, and 120 steps for dengue, chikungunya, and Zika. Metrics include F1-score and Area Under the ROC Curve (AUC-ROC) for each disease and phase. Bold values highlight the best result within each disease and metric.}
\begin{tabular}{lcc|cc|cc}
\toprule
\textbf{Steps} & \multicolumn{2}{c|}{\textbf{Dengue}} & \multicolumn{2}{c|}{\textbf{Chikungunya}} & \multicolumn{2}{c}{\textbf{Zika}} \\ \midrule
 & \(\mathbf{F1Score}\) & \(\mathbf{AUC-ROC}\) & \(\mathbf{F1Score}\) & \(\mathbf{AUC-ROC}\) & \(\mathbf{F1Score}\) & \(\mathbf{AUC-ROC}\) \\ \midrule
\multicolumn{7}{l}{\textbf{Validation}} \\ 
60 & 0.80 & 0.99 & \textbf{0.77} & \textbf{0.98} & 0.77 & 0.98 \\
90 & \textbf{0.94} & \textbf{0.99} & 0.76 & 0.97 & 0.81 & 0.97 \\
120 & 0.93 & 0.99 & 0.74 & 0.96 & \textbf{0.87} & \textbf{0.98} \\ \midrule
\multicolumn{7}{l}{\textbf{Testing}} \\ 
60 & \textbf{0.75} & \textbf{0.96} & 0.34 & 0.97 & \textbf{0.87} & 0.87 \\
90 & 0.59 & 0.93 & 0.33 & 0.97 & 0.86 & \textbf{0.90} \\
120 & 0.70 & 0.94 & \textbf{0.46} & \textbf{0.97} & 0.83 & 0.88 \\ \bottomrule
\end{tabular}
\label{table:f1_and_auc}
\end{table*}

%% file: section/results.tex
\section{Results}\label{results}

This section presents the results of the multitask LSTM model for predicting outbreaks and case counts of dengue, chikungunya, and Zika. Models were evaluated using input window sizes of 60, 90, and 120 days. Regression performance was measured using the Median Absolute Percentage Error (MedAPE) and Mean Absolute Percentage Error (MAPE), while classification was assessed using F1-score and Area Under the ROC Curve (AUC).

\subsection{Regression Performance}

Table~\ref{table:forecast_erros} summarizes regression errors across diseases and input window sizes. Dengue forecasts consistently yielded the lowest errors, with the Simple LSTM achieving optimal results at 120 steps (MedAPE: 4.45\%, MAPE: 7.5\%). Chikungunya presented higher variability; the best MedAPE (5.7\%) was observed at 60 steps using the Simple LSTM, while the lowest error using Bidirectional LSTM occurred at 120 steps. Zika predictions showed the highest overall error, reflecting the limited data available, though the Simple LSTM still performed best at 120 steps (MedAPE: 7.5

These results indicate that longer input sequences improve regression performance for dengue and Zika, while chikungunya benefits from shorter windows. The Simple LSTM outperformed the Bidirectional model in most cases.

\subsection{Classification Performance}

Table~\ref{table:f1_and_auc} presents the classification metrics. During validation, the best F1-score for dengue was 0.94 using a 90-day window, with an AUC of 0.99. However, when tested on unseen 2023 data, performance declined: the F1-score dropped to 0.59 and AUC to 0.93. Despite this, models with 60- and 120-day inputs maintained AUCs above 0.94, suggesting generalization for outbreak detection.

For chikungunya and Zika, the model achieved stable validation metrics but demonstrated sensitivity to test conditions, particularly for chikungunya (lowest F1: 0.33 at 90 steps). Zika results were more consistent, with the highest testing F1-score (0.87) obtained using a 60-step input.

Figure~\ref{fig:95CI} presents the 95\% bootstrap confidence intervals for F1 and AUC metrics. These intervals confirm the general reliability of the model, while highlighting sensitivity to input window size and disease characteristics.

\subsection{Summary of Findings}

The results suggest that the multitask LSTM model provides reliable predictions across both classification and regression tasks. Performance varies by disease and sequence length, suggesting the need for task-specific tuning. Dengue forecasting benefited from longer sequences, while outbreak classification peaked at intermediate windows. Zika demonstrated stable behavior despite data scarcity, and the Simple LSTM generally outperformed the Bidirectional variant. These insights reinforce the effectiveness of multitask learning for integrated epidemiological forecasting.

%% file: section/conclusion.tex
\section{Conclusion}\label{conclusion}

This paper investigated a multitask Long Short-Term Memory (LSTM) model for predicting arboviral outbreaks in Recife, Brazil, using publicly available epidemiological data. The proposed architecture jointly addressed outbreak detection and case forecasting for dengue, chikungunya, and Zika, enabling a unified assessment of temporal disease patterns.

Empirical results demonstrated that input window length plays a critical role in model performance. For regression tasks, longer sequences (120 steps) yielded improved accuracy, particularly for dengue. Classification performance was more sensitive to sequence length and disease type, with optimal F1-scores observed at intermediate windows (90 steps). Despite performance degradation on unseen test data—a common challenge in time series forecasting—the model maintained acceptable generalization across tasks.

The findings underscore the potential of multitask deep learning frameworks in supporting data-driven public health surveillance, especially in resource-constrained settings. Using a single architecture for multiple objectives, the model offers a scalable and efficient approach to outbreak monitoring.

Future research will explore the integration of exogenous variables such as climate, mobility, and socioeconomic indicators to enhance predictive power and applicability to other geographic and epidemiological contexts. Comparisons with single-task baselines and alternative neural architectures will also be pursued to further validate the benefits of multitask learning in infectious disease modeling.